%% file: bare_jrnl.tex
\documentclass[journal]{IEEEtran}
\usepackage{cite}

\usepackage[pdftex]{graphicx}
\pdfoutput=1

\ifCLASSINFOpdf
\else
\fi
\usepackage{amsmath}
\usepackage{array}
\usepackage[utf8]{inputenc} 
\usepackage[T1]{fontenc}    
\usepackage{hyperref}       
\usepackage{url}            
\usepackage{booktabs}       
\usepackage{amsfonts}       
\usepackage{nicefrac}       
\usepackage{microtype}      

\usepackage{graphicx}
\usepackage{subcaption}
\usepackage{booktabs} 

\usepackage{amsmath}
\usepackage{amssymb}
\usepackage{amsthm}
\usepackage{stmaryrd}
\usepackage{xcolor}
\usepackage{todonotes}

\usepackage[english]{babel} 
\usepackage{flexisym}
\usepackage{multirow}

\usepackage{enumitem}

\usepackage[noend]{algpseudocode}
\usepackage{algorithm2e}

\usepackage{etoolbox}
\newtoggle{supplementary}
\toggletrue{supplementary}

\begin{document}
%
\title{Deep K-SVD Denoising}
%
%
%

\author{Meyer~Scetbon,
Michael~Elad,~\IEEEmembership{Fellow,~IEEE,}
        and~Peyman~Milanfar,~\IEEEmembership{Fellow,~IEEE}
\thanks{M. Scetbon is with CREST - ENSAE , \texttt{meyer.scetbon@ensae.fr}.  
M. Elad and P. Milanfar are with Google Research, \texttt{[melad,milanfar]@google.com}.}}

%
%

\markboth{Journal of \LaTeX\ Class Files,~Vol.~14, No.~8, August~2015}%
{Shell \MakeLowercase{\textit{et al.}}: Bare Demo of IEEEtran.cls for IEEE Journals}
%



\maketitle

\begin{abstract}
This work considers noise removal from images, focusing on the well known K-SVD denoising algorithm. This sparsity-based method was proposed in 2006, and for a short while it was considered as state-of-the-art. However, over the years it has been surpassed by other methods, including the recent deep-learning-based newcomers. The question we address in this paper is whether K-SVD was brought to its peak in its original conception, or whether it can be made competitive again. The approach we take in answering this question is to redesign the algorithm to operate in a supervised manner. More specifically, we propose an end-to-end deep architecture with the exact K-SVD computational path, and train it for optimized denoising. Our work shows how to overcome difficulties arising in turning the K-SVD scheme into a differentiable, and thus learnable, machine. With a small number of parameters to learn and while preserving the  original K-SVD essence, the proposed architecture is shown to outperform the classical K-SVD algorithm substantially, and getting closer to recent state-of-the-art learning-based denoising methods. Adopting a broader context, this work touches on themes around the design of deep-learning solutions for image processing tasks, while paving a bridge between classic methods and novel deep-learning-based ones. 
\end{abstract}


%
\IEEEpeerreviewmaketitle

\input{1-Intro}

\input{2-K-SVD}
\input{3-OurMethod}
\input{4-Results}

\input{5-Conclu}

\ifCLASSOPTIONcaptionsoff
  \newpage
\fi



%
\bibliographystyle{abbrv}
\bibliography{6-Biblio}




%








\end{document}

%% file: 1-Intro.tex
\section{Introduction}

\IEEEPARstart{T}{his} paper addresses the classic image denoising problem: an ideal image $\mathbf{x}$ is measured in the presence of an additive zero-mean white and homogeneous Gaussian noise, $\mathbf{v}$, with standard deviation $\sigma$. The measured image $\mathbf{y}$ is thus $\mathbf{y}=\mathbf{x}+\mathbf{v}$, and our goal is the recovery of $\mathbf{x}$ from $\mathbf{y}$ with the knowledge of the parameter $\sigma$. This is quite a challenging task due to the need to preserve the fine details in $\mathbf{x}$ while rejecting as much noise as possible.    

The importance of the image denoising problem cannot be overstated. First and foremost, noise corruption is inevitable in any image sensing process, often times heavily degrading the visual quality of the acquired image. Indeed, today's cell-phones all deploy a denoising algorithm of some sort in their camera pipelines \cite{plotz2017benchmarking}. Removing noise from an image is also an essential and popular pre-step in various image processing and computer vision tasks \cite{katsaggelos2012digital}. Last but not least, many image restoration problems can be addressed effectively by solving a series of denoising sub-problems, further broadening the applicability of image denoising algorithms \cite{afonso2010fast,romano2017little}. Due to its practical importance and the fact that it is the simplest inverse problem, image denoising has become the entry point for many new ideas brought over the years to the realm of image processing. Over a period of several decades, many image denoising algorithms have been proposed and tested, forming an evolution of methods with gradually improved performance. 

A common and systematic approach for the design of novel denoising algorithms is the Bayesian point of view. This calls for image priors, used as regularizers within the Maximum a Posteriori (MAP) or the Minimum Mean Squared Error (MMSE) estimators. In this paper we concentrate on one specific regularization approach, as introduced in \cite{elad2006image}: the use of sparse and redundant representation modeling of image patches -- this is the K-SVD denoising algorithm, which stands at the center of this paper. The authors of \cite{elad2006image} defined a global image prior that forces sparsity over patches in every location in the image. Their algorithm starts by breaking the image into small fully overlapping patches, solving their MAP estimate (i.e., finding their sparse representation), and ending with a tiling of the results back together by an averaging. As the MAP estimate relies on the availability of the dictionary, this work proposed two approaches, both harnessing the well known K-SVD dictionary learning algorithm \cite{aharon2006k}. The first option is to train off-line on an external large corpus of image patches, aiming for a universally good dictionary to serve all test images. The alternative, which was found to be more effective, suggests using the noisy patches themselves in order to learn the dictionary, this way adapting to the denoised image. 

K-SVD has been widely used and extended, as evidenced by its many followup papers. For a short while, this algorithm was considered as state-of-the-art, standing at the top in denoising performance\footnote{Ranking denoising algorithms is typically done by evaluating synthetic denoising performance on agreed-upon image databases (e.g. set12 or BSD68), measuring Peak-Signal-to-Noise (PSNR) and/or Structured Similarity Index Measure (SSIM) results.}. However, over the years it has been surpassed by other methods, such as BM3D \cite{dabov2007video}, EPLL \cite{zoran2011learning}, WNNM \cite{gu2014weighted}, and many others. The recent newcomers to this game -- supervised deep-learning based denoising methods -- are currently at the lead \cite{chen2016trainable,lefkimmiatis2017non, zhang2017beyond,liu2018non,zhang2018ffdnet}. 

Can K-SVD denoising make a comeback and compete favorably with the most recent and best performing denoising algorithms? In this paper we answer this question positively. We aim to show that the K-SVD denoising algorithm can be brought to perform far better by considering a different training strategy. This leads to far better results which should be taken as reference when comparing new methods with the K-SVD algorithm. While its original version trained a dictionary for getting sparse representations, this new end-to-end training outperforms it provided that its parameters are tuned in a supervised manner. By following the exact K-SVD computational path, we preserve its global image prior. This includes (i) breaking the image into small fully overlapping patches, (ii) solving their MAP estimate as a pursuit that aims to get their sparse representation in a learned dictionary, and then (ii) averaging the overlapping patches to restore the clean image. A special care is given to the redesign of all these steps into a differentiable and learnable computational scheme. We therefore end up with a deep architecture that reproduces the exact K-SVD operations, and can be trained by back-propagation for best denoising results. Our work shows that with small number of parameters to learn and while preserving the original K-SVD essence, the proposed machine outperforms the original K-SVD and other classical algorithms (e.g. BM3D and WNNM), and getting closer to state-of-the-art learning based denoising methods. 

We should note that the performance of the proposed method still falls short when compared to leading state-of-the-art deep learning based denoising methods. Further work is required to close this gap, and we outline options for this feat in the discussion towards the end of the paper. However, we emphasize that it is not the goal of our paper to propose yet another denoiser with top performance. Rather, our prime goal in this work is to offer an appealing bridge between classical methods in image processing and the new era of deep neural networks, with the hope to pave the way to followup work that will show the synergy that could exist between the two paradigms. We contribute to the construction of this bridge by focusing on the K-SVD denoising algorithm, showing that it can be treated as a deep network and trained as such, and demonstrating the fact that it can be substantially improved via supervised learning.

Alongside the obtained boost in performance, compared to the original method we embark from, the resulting network has other valuable benefits. The obtained network has a clear and meaningful interpretation of its parameters and the data flowing in it, a property very much lacking in other deep denoisers, and something that could be leveraged in various ways. Please recall that the rationale of sparse decomposition of signals has been found useful in many other applications in the field of compressed sensing and signal recovery. 

Another benefit has to do with the fact that the network obtained is more concise, implying that it can be trained with less data, or even adapted to the incoming image, as in \cite{LIDIA}. Beyond all these, and more importantly, the main message we aim to convey is that design of neural architectures for image processing tasks could be done differently, replacing the arbitrary trial-and-error networks by well-motivated ones that emerge from the classic know-how. All the above represents an ambitious endeavor, and our work offers few of the first steps in this long track of required work.

More specifically, by rewriting the chain of operations in the K-SVD algorithm in a differentiable manner, we are able to back-propagate through its parameters and obtain an algorithm which performs much better than all its earlier variants. Indeed, the resulting network is almost consistently better than all other model-based methods and some widely used deep learning based methods as well. Please note that our strategy embarks from the weaker version of the K-SVD denoising algorithm that relies on a universal dictionary for all images (as opposed to the image-adapted option), and yet it shows that by a task-driven design of it's parameters, a much better suited \emph{dictionary} and overall denoising performance are within reach. 


This paper is organized as followed. Section \ref{section-KSVD} recalls the K-SVD denoising algorithm, serving as the background for our derived alternative. In Section \ref{section-OurMethod} we present the designed architecture with various modifications and adjustments that enable differentiabilty, local adaptivity, and more. Section \ref{section-Results} describes series of experiments that demonstrate the superiority of the proposed learned network over the classic K-SVD denoising algorithm, and shows the tendency of our proposed network to have competitive performance with recent learned methods. We conclude this work in Section \ref{section-Discussion} with a wide discussion about this work and its contributions, and highlight potential future research directions.

%% file: 2-K-SVD.tex
\section{The K-SVD Denoising Algorithm}
\label{section-KSVD}

In \cite{elad2006image} the authors address the image denoising problem by using local sparsity and redundancy as ingredients in the formation of a global Bayesian objective. In this section we describe this K-SVD denoising algorithm by discussing (i) their global prior; (ii) the objective function induced; (iii) its corresponding numerical solver; and (iv) the two approaches for training the corresponding dictionary.


\subsection{From the Patch- to a Global Objective Function}

We start by introducing the local prior as imposed on patches in \cite{elad2006image}. Let $\mathbf{x}$ be a small image patch of size $\sqrt{p}\times \sqrt{p}$ pixels, ordered lexicographically as a column vector of length $p$. The sparse representation model assumes that $\mathbf{x}$ is built as a linear combination of $s \ll p$ columns (also referred to as atoms) taken from a pre-specified dictionary\footnote{The option $m>n$ implies that the dictionary is redundant.} $\mathbf{D}\in\mathbb{R}^{p\times m}$. Put formally, $\mathbf{x}=\mathbf{D}\alpha$, where $\alpha \in \mathbb{R}^m$ is a sparse vector with $s$ non-zeros (this is denoted by $\Vert \alpha\Vert_0=s$). Consider $\mathbf{y}$, a noisy version of $\mathbf{x}$, contaminated by an additive zero-mean white Gaussian noise with standard deviation $\sigma$. The MAP estimator for denoising this patch is obtained by solving
\begin{equation}
\label{sparse-code-constr}
\hat{\alpha} = \arg\min_{\alpha}~~\Vert \alpha\Vert_0 \quad \text{s.t.}\quad 
\Vert \mathbf{D}\alpha - \mathbf{y} \Vert_2^2 \leq p \sigma^2, 
\end{equation}
aiming to recover the sparse representation vector of $\mathbf{x}$. This is followed by $\hat{\mathbf{x}}=\mathbf{D}\hat{\alpha}$, obtaining the denoised result \cite{chen2001atomic, donoho2005stable,tropp2006just}. Note that the above optimization can be changed to a Lagrangian form, 
\begin{equation}
\label{sparse-code-regul}
\hat{\alpha} = \arg\min_{\alpha} ~~\lambda \Vert \alpha\Vert_0 + \frac{1}{2} \Vert \mathbf{D}\alpha - \mathbf{y} \Vert_2^2, 
\end{equation}
such that the constraint becomes a penalty. With a proper choice of $\lambda$, which is signal (the vector $\mathbf{y}$) dependent, the two problems can become equivalent. 

Moving now to handle a complete and large image $\mathbf{X}$ of size $\sqrt{N}\times \sqrt{N}$ and its noisy version $\mathbf{Y}$ (both held as vectors of length $N$), the global image prior proposed in \cite{elad2006image} imposes the above-described local prior on every patch in $\mathbf{X}$, considering their extractions with full overlaps. This leads to the following global MAP estimator for the denoising:
\begin{equation}
\begin{aligned}
\label{global-objective}
&\min_{\{\alpha_{k}\}_k,\mathbf{X}}~~ \frac{\mu}{2} \Vert \mathbf{X}-\mathbf{Y}\Vert_2^2
+\\
&\sum_{k}\left(\lambda_{k}\Vert \alpha_{k}\Vert_0 + \frac{1}{2}\Vert \mathbf{D}\alpha_{k} - \mathbf{R}_{k}\mathbf{X} \Vert_2^2\right). 
\end{aligned}
\end{equation}

In this expression, the first term is the log-likelihood global force that demands a proximity between the measured image, $\mathbf{Y}$, and its denoised (and unknown) version $\mathbf{X}$. Put as a constraint, this penalty would have read $\Vert \mathbf{X}-\mathbf{Y}\Vert_2^2\leq N\sigma^2$, which reflects the direct relationship between $\mu$ and $\sigma$.

The second term stands for the image prior that assures that in the constructed image, $\mathbf{X}$, every patch\footnote{For simplicity and without loss of generality, a single index is used to account for the spatial image location.}  $\mathbf{x}_{k}=\mathbf{R}_{k}\mathbf{X}$ of size $\sqrt{p}\times \sqrt{p}$ in every location (thus, the summation by $k$) has a sparse representation with bounded error. The matrix $\mathbf{R}_{k} \in \mathbb{R}^{p\times N}$ stands for an operator that extracts the $k$-th block from the image. 
As to the coefficients $\lambda_{k}$, those must be spatially dependent, so as to comply with a set of constraints of the form  $\Vert \mathbf{D}\alpha_{k} - \mathbf{x}_{k} \Vert_2^2\leq p\sigma^2$.


\subsection{Numerical Solution}

Assume for the moment that the underlying dictionary $\mathbf{D}$ is known. The objective function in Equation (\ref{global-objective}) has two kinds of unknowns: the sparse representations $\alpha_{k}$ per each location, and the output image $\mathbf{X}$. Instead of addressing both together, the authors of \cite{elad2006image} propose a block-coordinate minimization algorithm that starts with an initialization $\mathbf{X}=\mathbf{Y}$, and then seeks the optimal $\hat{\alpha}_{k}$ for all locations $k$. This leads to a decoupling of the minimization task to many smaller pursuit problems of the form
\begin{align}
\label{obj-sparse}
\hat{\alpha}_{k}=\arg\min_{\alpha_{k}}~~\lambda_{k}\Vert \alpha_{k}\Vert_0 +  \frac{1}{2} \Vert \mathbf{D}\alpha_{k} - \mathbf{x}_{k} \Vert_2^2,
\end{align}
each handling a separate patch. This is solved in \cite{elad2006image} using the Orthonormal Matching Pursuit (OMP) \cite{elad2010sparse}, which gathers one atom at a time to the solution, and stops when the error $\Vert \mathbf{D}\alpha_{k} - \mathbf{x}_{k} \Vert_2^2$ goes below\footnote{In fact, the threshold used in \cite{elad2006image} is $c \cdot p\sigma^2$, with $c=1.15$, which was found empirically to perform best.}  $p\sigma^2$. This way, the choice of $\lambda_{k}$ has been handled implicitly. Thus, this stage works as a sliding window sparse coding stage, operated on each patch of size $\sqrt{p}\times \sqrt{p}$ pixels at a time. 

Given all the sparse representations of the patches, $\{\hat \alpha_{k}\}_k$, we can now fix those and turn to update $\mathbf{X}$. Returning to the expression in Equation (\ref{global-objective}), we need to solve
\begin{align}
\hat{\mathbf{X}}= \arg\min_{\mathbf{X}}~~  \frac{\mu}{2} \Vert \mathbf{X}-\mathbf{Y}\Vert_2^2 +  \frac{1}{2}\sum_{k} \Vert \mathbf{D}\hat{\alpha}_{k} - \mathbf{R}_{k}\mathbf{X} \Vert_2^2.
\end{align}
This is a simple quadratic term that has a closed-form solution of the form
\begin{align}
\label{global-image}
\hat{\mathbf{X}}=\left(\sum_{k}\mathbf{R}_{k}^{T}\mathbf{R}_{k}+ \mu\mathbf{I}\right)^{-1}\left(\mu\mathbf{Y}+\sum_{k}\mathbf{R}_{k}^{T}\mathbf{D}\hat{{\alpha}}_{k}\right).
\end{align}
The matrix to invert in the above expression is a diagonal one, and thus the required computation is quite simple. In fact, all that this expression does is to put back the patches to their original locations, and average these with a weighted version of the noisy image itself. 

All the above stands for a single update of $\{\alpha_{k}\}_k$ and then $\hat{\mathbf{X}}$. For an effective block-coordinate minimization of the cost function in Equation (\ref{global-objective}) we should repeat these pair of updates several times. However, a difficulty with such an approach is the fact that once $\hat{\mathbf{X}}$ has been modified, we no longer know the level of noise in each patch, and thus the stopping criteria for the OMP becomes more challenging. The original K-SVD denoising algorithm, as proposed in \cite{elad2006image}, chose to apply only the first round of updates. The work reported in \cite{sulam2015expected} adopts an EPLL point of view \cite{zoran2011learning}, extending the iterative algorithm further for getting improved results. 


\subsection{Obtaining the Dictionary $\mathbf{D}$}

The discussion so far has been based on the assumption that the dictionary $\mathbf{D}$ is known. This could be the case if we train it using the K-SVD algorithm over a corpus of  clean image patches \cite{elad2010sparse}. An interesting alternative is to embed the identification of $\mathbf{D}$ within the Bayesian formulation. Returning to the objective function in Eq. (\ref{global-objective}), the authors of \cite{elad2006image} also considered the case where $\mathbf{D}$ is an unknown, 
\begin{align}
\min_{\{\alpha_{k}\}_k,\mathbf{X},\mathbf{D}} ~~ \frac{\mu}{2} \Vert \mathbf{X}-\mathbf{Y}\Vert_2^2 +  \sum_{k} \left(\lambda_{k}\Vert \alpha_{k}\Vert_0 + \frac{1}{2} \Vert \mathbf{D}\alpha_{k} - \mathbf{R}_{k}\mathbf{X} \Vert_2^2 \right). \nonumber
\end{align}
In this case, $\mathbf{D}$ is learned using all the existing noisy patches taken from $\mathbf{Y}$ itself. Put more formally, a block-coordinate minimization is done: Initialize the dictionary $\mathbf{D}$ as the overcomplete DCT matrix and set $\mathbf{X} = \mathbf{Y}$. Then iterate between the OMP over all the patches and an update of $\mathbf{D}$ using the K-SVD strategy \cite{aharon2006k}. After $T=10$ such rounds, the dictionary admits a content adapted to the image being treated, and the representations $\{\hat \alpha_{k}\}_k$ are ready for a final stage in which the output image is computed via Eq. (\ref{global-image}).


%% file: 3-OurMethod.tex
\section{Proposed Architecture}
\label{section-OurMethod}


In this work our goal is to design a network that reproduces the K-SVD denoising algorithm, while having the capacity to better learn its parameters. By reposing each of the operations within the K-SVD algorithm in a differentiable manner, we aim to be able to back-propagate through its parameters and obtain a version of the algorithm that outperforms its earlier variants. Note that in this supervised mode of work, we adopt the weaker version of the K-SVD denoising algorithm that relies on a universal dictionary for all images, as opposed to the image-adapted option that was shown to be superior in \cite{elad2006image}.

One of the main difficulties we encounter is the pursuit stage, in which we are supposed to replace the greedy OMP algorithm by an equivalent learnable alternative. This may seem as an easy task, as we can use the $L_1$-based Iterated Soft-Thresholding Algorithm (ISTA), unfolded appropriately for several iterations \cite{gregor2010learning, daubechies2004iterative}. However, the challenge is the fact that OMP easily adapts the treatment for each patch using a stopping criterion based on the noise level. The equivalence in the ISTA case requires an identification of the appropriate regularization parameter $\lambda_k$ for each patch, which is a non-trivial task. 

Assuming that this issue has been resolved, our computational process includes a decomposition of the image into its overlapped patches, cleaning of each by an appropriate pursuit, and a reconstruction of the overall image by averaging the cleaned patches. We propose to learn the parameters of this network by training over pairs of corrupted and ground-truth images. Next, we describe in details this overall architecture.


\subsection{Patch Denoising}

\begin{figure*}[!ht]
\centering
\includegraphics[width=0.8\textwidth]{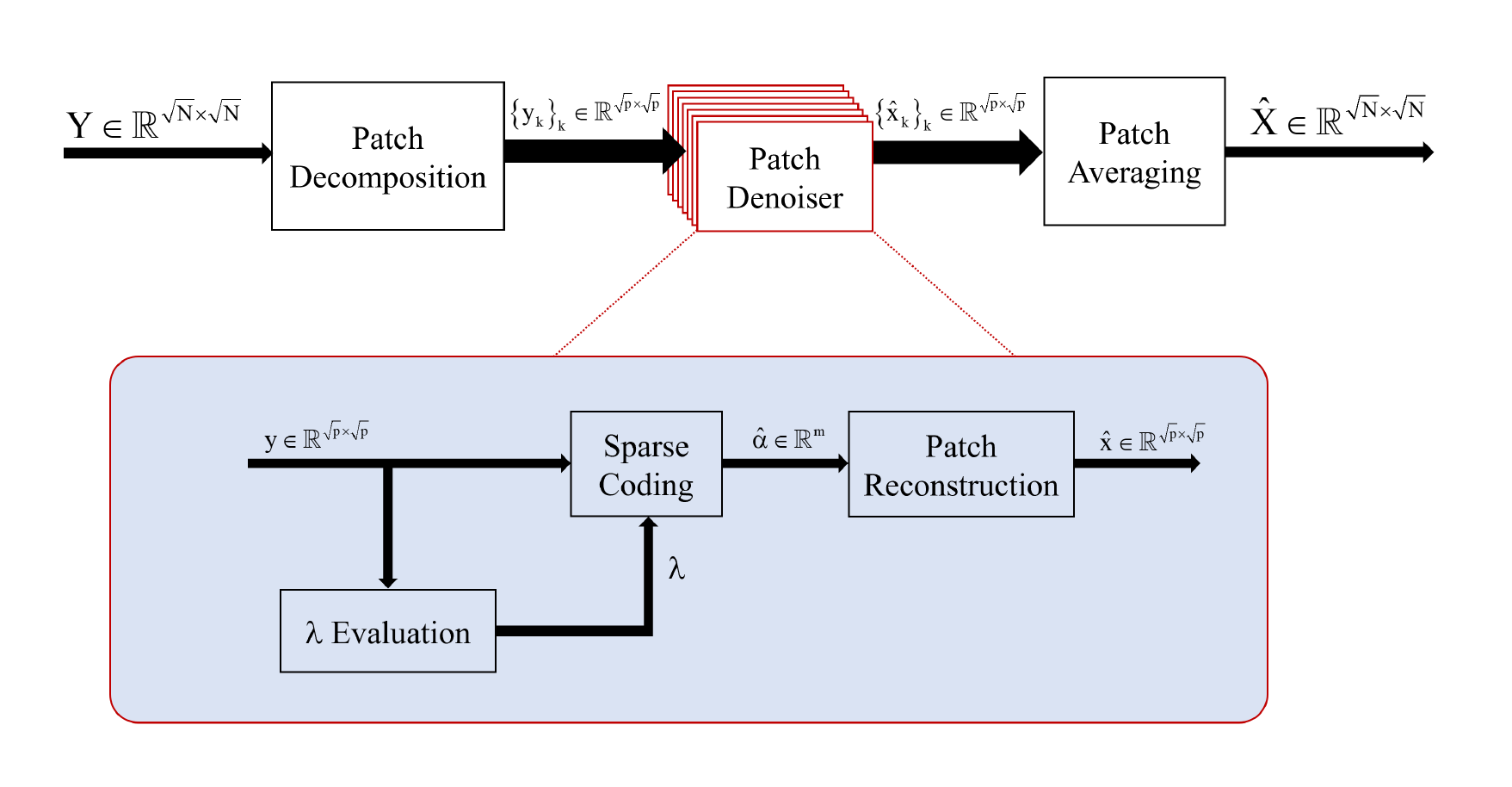}
\caption{Architecture of the proposed method.\label{proposed}}
\end{figure*}

Figure \ref{proposed} illustrates our end-to-end architecture. We start by describing the three stages that perform the denoising of the individual patches.

\textbf{Sparse Coding:} Given a patch $\mathbf{y}\in\mathbb{R}^{\sqrt{p}\times \sqrt{p}}$ (held as a column vector of length $p$) corrupted by an additive zero-mean Gaussian noise with standard deviation $\sigma$, we aim to derive its sparse code according to a known dictionary $\mathbf{D}\in\mathbb{R}^{p\times m}$. This objective can be formulated as in Equation  (\ref{sparse-code-constr}). An approximate solution to this problem can be obtained by replacing the $\ell_0$-norm with an $\ell_1$ \cite{donoho2003optimally, donoho2005stable}:
\begin{equation}
\label{l1-constraint}
\min_{\alpha\in\mathbb{R}^m}\Vert \alpha\Vert_1 \quad \text{s.t}\quad 
\Vert \mathbf{D}\alpha - \mathbf{y} \Vert_2^2 \leq p\sigma^2.
\end{equation}
For a proper choice of $\lambda$, the above  can be reformulated as
\begin{equation}
\label{l1-penalty}
\hat{\alpha} = \arg\min_{\alpha} \frac{1}{2}\Vert \mathbf{D}\alpha - \mathbf{y}\Vert_2^2 + \lambda \Vert \alpha \Vert_1. 
\end{equation}
A popular and effective algorithm for solving the above problem is the Iterative Soft Thresholding Algorithm (ISTA) \cite{daubechies2004iterative}, which is guaranteed to converge to the global optimum
\begin{align}
\label{recur-ista}
\hat{\alpha}_{t+1} = S_{\lambda/c}\left(\hat{\alpha_t} - \frac{1}{c}\mathbf{D}^{T}(\mathbf{D}\hat{\alpha}_t - \mathbf{y})\right)\quad;\quad \hat{\alpha}_{0}=0, 
\end{align}
where $c$ is the square spectral norm of $\mathbf{D}$ and $S_{\lambda/c}$ is the component-wise soft-thresholding operator,
\begin{align}
[S_{\theta}(\mathbf{v})]_i=\text{sign}(v_i)(\vert v_i\vert - \theta)_{+}.
\end{align}
The motivation to adopt a proximal gradient descent method, as done above, is the fact that it allows an unrolling of the sparse coding stage into a meaningful and learnable scheme, just as practiced in \cite{gregor2010learning}. Indeed, replacing the $\ell_0$-norm by the $\ell_1$ supports this goal as it allows to differentiate through this scheme. Moreover the iterative formula given by Eq.~(\ref{recur-ista}) is operated on each patch, which means that it is just like a \emph{convolution} in terms of operating on the whole image. Because of these reasons, in this work we consider a learnable version of ISTA  by keeping exactly the same recursion with a fixed number of iterations $T$, and letting $c$ and $\mathbf{D}$ become the learnable parameters. Note that we also consider the $\ell_0$-based proximal gradient descent where the soft-thresholding is replaced by an hard one which leads to worst results. The smoothness given by the $\ell_1$-norm gives a more dense dictionary which deals better to remove the noise.

\textbf{$\lambda$ Evaluation:} Referring to the pursuit formulation in Equation (\ref{l1-penalty}), an important issue is the need to set the parameter $\lambda$. This regularization coefficient depends not only on $\sigma$ but also on the patch $\mathbf{y}$ itself. Following the computational path of the K-SVD denoising algorithm in \cite{elad2006image}, we should set $\lambda_k$ for each patch $\mathbf{y}_k$ so as to yield sparse representation with a controlled level of error, $\Vert \mathbf{D}\hat{\alpha}_k - \mathbf{y}_k \Vert_2^2 \leq p\sigma^2$. As there is no closed-form solution to this evaluation of $\lambda$-s, we propose to learn a regression function from the patches $\mathbf{y}_{k}$ to their corresponding regularization parameters $\lambda_{k}$. A Multi-Layer Perceptron (MLP) network is used to represent this function, $\lambda = f_{\theta}(\mathbf{y})$, where $\theta$ is the vector of the parameters of the MLP. Our MLP consists of three hidden layers, each composed of a fully connected linear mapping followed by a ReLU (apart from the last layer). The input layer has $p$ nodes, which is the dimension of the vectorized patch, and the output layer consists of a single node, being the regularization parameter. The overall structure of the network is given by the following expression, in which $[a \times b]$ symbolizes a multiplication by a matrix of that size: $\mbox{MLP:}~~ \mathbf{y}   \rightarrow  \left[ p \times 2p \right] \rightarrow \mbox{ReLU} \rightarrow \left[ 2p \times p \right] \rightarrow \mbox{ReLU} \rightarrow \left[ p/2 \times 1 \right] \rightarrow \lambda$. Thus, an overall of nearly $4p^2$ parameters are needed for this regression network. 


\textbf{Patch Reconstruction}: This stage reconstructs the cleaned version $\hat {\mathbf{x}}$ of the patch $\mathbf{y}$ using $\mathbf{D}$ and the sparse code $\hat \alpha$. This is given by $\hat{\mathbf{x}}=\mathbf{D}\hat{\alpha}$. Note that in our learned network, the dictionary stands for a set of parameters that are shared in all locations where we multiply by either $\mathbf{D}$ or $\mathbf{D}^T$. The proposed approach based on the patch-decomposition and reconstruction is closely related to a convolutional sparse coding approach as it has been shown in~\cite{NIPS2019_8499}.


\subsection{End-to-end Architecture}

We can now discuss the complete architecture. We start by breaking the input image into fully overlapping patches, then treat each corrupted patch via the above-described patch denoising stage, and conclude by rebuilding the image by averaging the cleaned version of these patches. In the last stage we slightly deviate from the original K-SVD, by allowing a learned weighted combination of the patches. Denoting by $\mathbf{w}\in\mathbb{R}^{\sqrt{p}\times \sqrt{p}}$ this patch of weights, the reconstructed image is obtained by 
\begin{align}
\hat{\mathbf{X}}=\frac{\sum\limits_{k} \mathbf{R}^{T}_{k}(\mathbf{w}\odot \hat{\mathbf{x}}_k)}{\sum\limits_{k}\mathbf{R}^{T}_{k}\mathbf{w}}, 
\end{align}
where $\odot$ is the Schur product, and the division is done element-wise. This weighted averaging aligns with Guleryuz' approach as advocated in \cite{guleryuz2007}.

To conclude, the proposed network $F$ is a parametrized function of $\theta$ (the parameters of the MLP network computing $\lambda$), $c$ (the step-size in the ISTA algorithm), $\mathbf{D}$ (the dictionary) and $\mathbf{w}$ (the weights for the patch-averaging). The overall number of parameters stands on $p(4p+m+3/2)+1$; for example, for $p=64$ and $m=256$, this number is $32,865$. 

Given a corrupted image $\mathbf{Y}$, the computation $\hat{\mathbf{X}}=F(\mathbf{Y})$ returns a cleaned version of it. Training $F$ is done by minimizing the loss function $\mathcal{L} =  \sum_i \Vert \mathbf{X}_i-F(\mathbf{Y}_i) ||_2^2$, with respect to all the above parameters. In the above objective, the set $\{\mathbf{X}_i \}_i$ stands for our training images, and $\{\mathbf{Y}_i \}_i$ are their synthetically noisy versions, obtained by $\mathbf{Y}_i=\mathbf{X}_i+\mathbf{V}_i$, where $\mathbf{V}_i$ is a zero mean and white Gaussian iid noise vector.



To better clarify, we note that our scheme does not propose an online training on the incoming image to be denoised. Rather, the strategy proposed is a differentiable approximation of the K-SVD operations, and training this network off-line once to set its parameters. Once completed, this universal denoiser is ready for use as a simple and fixed inference of the given network on any incoming image.


\subsection{Extension to Multiple Update}

As already mentioned in the previous section, an EPLL version of the K-SVD can be envisioned, in which the process of cleaning the patches is repeated several times. This implies that once the above architecture obtains its output $\hat{\mathbf{X}}$, the whole scheme could be applied again (and again). This diffusion process of repeated denoisings has been shown in \cite{sulam2015expected} to improve the K-SVD denoising performance. However, the difficulty is in setting the noise level to target in each patch after the first denoising, as it is no longer $p\sigma^2$. In our case, we adopt a crude version of the EPLL scheme, in which we disregard the noise level problem altogether, and simply assume that the $\lambda $ evaluation stage takes care of this challenge, adjusting the MLP in each round to best predict the $\lambda$ values to be used. Thus, our iterated scheme shares the dictionary across all denoising stages, while allowing a different $\lambda$ evaluation network for each stage.  

%% file: 4-Results.tex
\section{Experimental Results} 
\label{section-Results}

We turn to present experiments with the proposed Learned K-SVD (LKSVD). Our goals are to show that LKSVD is
\begin{itemize}
    \item Much better than the original KSVD in its two forms  -- the image adaptive algorithm ($\mbox{KSVD}_1$), and the one using a universal dictionary ($\mbox{KSVD}_2$); 
    \item Better than other classic denoising algorithms; and  
    \item Competitive with recent deep-learning based denoisers. 
\end{itemize}


\subsection{Training}

\textbf{Dataset:} In order to train our model we generate the training data using the Berkeley segmentation dataset (BSDS) \cite{MartinFTM01}, which consists of 500 images. We split these images into a training set of 432 images and the validation/test set that consists of the remaining 68 images. We note that these 68 images are exactly the ones used in the standard evaluation dataset of \cite{roth2009fields}. In addition, following \cite{liu2018non, zhang2017beyond}, we test our proposed method on the benchmark Set12 -- a collection of widely-used testing images. The training and the two test sets are strictly disjoint and all the images are converted to gray-scale in each experiment setup. This allows a fair and comprehensive comparison with recent deep learning based methods, as we train and test on the same datasets and benchmarks used in \cite{lefkimmiatis2017non, lefkimmiatis2018universal, zhang2017beyond, liu2018non, chen2016trainable, mao2016image}. 


\textbf{Training Settings:} During training we randomly sample cropped images of size $128\times128$ from the training set. We add i.i.d. Gaussian noise with zero mean and a specified level of noise $\sigma$ to each cropped image as the noisy input during training. We train a different model for each noise level, considering $\sigma=15,25,50$.

We use SGD optimizer to minimize the loss function. We set the learning rate as $1e-4$ and consider one cropped image as the minibatch size during training. We use the same initialization as in the K-SVD algorithm to initialize the dictionary $\mathbf{D}$, i.e the overcomplete DCT matrix. We also initialize the normalization paramater $c$ of the sparse coding stage using the squared spectral norm of the DCT matrix. The other parameters of the network are randomly initialized using  Kaiming Uniform method. Training a model takes few hours with a Titan Xp GPU. 

\textbf{Test Settings:} Our network does not depend on the input size of the image. Thus, in order to test our architecture's performance, we simply add white Gaussian noise with a specified power to the original image, and feed it to the learned scheme. The metric used to determine the quality is the standard Peak-Signal-to-Noise (PSNR).


\subsection{Denoising Performance}

In Tables \ref{table-BSD}, \ref{table-BSD-SSIM} and \ref{table-Set12} we compare\footnote{The results in these tables corresponding to BM3D and WNNM have been taken from \cite{liu2018non} and \cite{zhang2018ffdnet}, respectively.} LKSVD with the two original K-SVD versions ($\mbox{KSVD}_1$ and $\mbox{KSVD}_2$) and two leading classic denoising algorithms, BM3D \cite{dabov2007video} and WNNM \cite{gu2014weighted}. Tables \ref{table-BSD} and \ref{table-BSD-SSIM} refer to the BSD68 test-set (one showing PSNR and the other SSIM quality measures) and Table \ref{table-Set12} shows the Set12 results (PSNR only). In this comparison, LKSVD is set to use the same patch and dictionary sizes as in $\mbox{KSVD}_1$ and $\mbox{KSVD}_2$ from \cite{elad2006image}, namely $p=64$ and $m=256$. Also, LKSVD applies $T=7$ unfolded iterations of ISTA, and $K=3$ EPLL-like denoising rounds.{\footnote{We have explored the case when the weights are not shared between the different EPLL-like updates, and we have observed a small improvement of $0.1-0.2$dB over the reported results. Thus, we omit this option in our presentation. }} 

\begin{table}[!h]
\centering
\scalebox{0.9}{
\begin{tabular}{ c | c | c c c c c  }
Dataset & Noise & BM3D & WNNM & $\mbox{KSVD}_1$ & $\mbox{KSVD}_2$ & LKSVD \\ 
\hline
\multirow{3}{*}{BSD 68}& 15 & 31.07 & 31.37 &  30.91  & 30.87 & \textbf{31.48}\\  
& 25 & 28.57 & 28.83 &  28.32 & 28.28 & \textbf{28.96}   \\  
& 50 & 25.62 & 25.87 & 25.03 &  25.01 &  \textbf{25.97} \\
\bottomrule
\end{tabular}}
\space
\caption{LKSVD vs. classic methods (BSD68): Denoising performance (PSNR [dB]) for various noise levels. \label{table-BSD}}
\end{table}

\begin{table}[!h]
\centering
\scalebox{0.9}{
\begin{tabular}{ c | c | c c c c c  }
Dataset & Noise & BM3D & WNNM & $\mbox{KSVD}_1$ & $\mbox{KSVD}_2$ & LKSVD \\ 
\hline
\multirow{3}{*}{BSD 68} & 15 &0.8717  & 0.8766  & 0.8692  & 0.8685 &\textbf{0.8835} \\  
 & 25  & 0.8013 & 0.8087 & 0.7876 &  0.7894 &  \textbf{0.8171} \\  
 & 50  & 0.6864 & 0.6982 & 0.6322  &  0.6462  &  \textbf{0.7035} \\
\bottomrule
\end{tabular}}
\space
\caption{LKSVD versus classic methods (BSD68): Denoising performance (SSIM) for various noise levels. \label{table-BSD-SSIM}}
\end{table}

\begin{table*}[!t]
\centering
\scalebox{1}{
\begin{tabular}{ c | c c c c c c c c c c c c | c }
Images  &C.man & Peppers & House & Airplane & Couple & Parrot & Man & Monarch & Starfish & Boat &  Barbara & Lena & Average \\ 
\hline
Noise level &  \multicolumn{12}{c}{$\sigma=15$} \\  
\hline
$\mbox{KSVD}_1$  & 31.43 & 32.21 & 34.23 & 30.80 & 31.59 & 30.99 & 31.64 & 31.45 & 30.95 & 31.83 & \textbf{32.44} & 33.78 & 31.95     \\  
$\mbox{KSVD}_2$  &  31.39 & 32.16 & 33.85  & 30.96 & 31.66  & 30.96 & 31.62 & 31.71 & 30.99 & 31.63 & 30.58 & 33.49 & 31.75 \\
LKSVD  &\textbf{32.16}  & \textbf{32.92} & 34.59 &\textbf{31.54} &\textbf{32.11} & \textbf{31.66} & \textbf{32.22} & \textbf{32.78} &\textbf{31.78} &\textbf{32.18}  & 32.22 & \textbf{34.24} & \textbf{32.53}  \\

\hline
Noise level &  \multicolumn{12}{c}{$\sigma=25$} \\  
\hline
$\mbox{KSVD}_1$ & 28.75 & 29.64 &31.86 & 28.21 & 29.10& 28.42 & 29.26 & 28.83 & 28.27 & 29.44 & \textbf{29.77} & 31.37 & 29.41\\
$\mbox{KSVD}_2$  & 28.78  & 29.74 & 31.46 & 28.39 & 29.04 & 28.57 & 29.16 & 28.85 & 28.24 & 29.18 & 27.61 & 31.04 & 29.17  \\
LKSVD   &\textbf{29.70}  &\textbf{30.35}  & \textbf{32.53} &\textbf{28.92} &\textbf{29.71}  & \textbf{29.13} & \textbf{29.85} &  \textbf{30.15} & \textbf{28.99} & \textbf{29.95} & 29.36 &  \textbf{31.99} & \textbf{30.05}\\ 

\hline
Noise level &  \multicolumn{12}{c}{$\sigma=50$} \\  
\hline
$\mbox{KSVD}_1$  &  25.12& 25.93 & 27.82 & 24.86 &25.56 & 24.80 & 26.16 & 25.11 & 24.45 & 25.98 & \textbf{25.78} & 27.71 & 25.78 \\  
$\mbox{KSVD}_2$  & 25.29 & 26.02 & 27.71 & 24.85 & 25.44 & 25.15 & 25.98 & 24.82 & 24.32 & 25.93 & 24.04 & 27.32 & 25.58  \\
LKSVD  &\textbf{26.68}& \textbf{26.96} & \textbf{29.37} & \textbf{25.62} & \textbf{26.55} & \textbf{25.99}& \textbf{26.95} &  \textbf{26.54} &\textbf{25.38} &\textbf{26.99} & 25.73 & \textbf{28.85} & \textbf{26.80} \\

\bottomrule
    \end{tabular}}
\space
\caption{LKSVD versus KSVD (Set12): Denoising performance (PSNR [dB]) for various noise levels. Best results are marked in bold.
\label{table-Set12}}
\end{table*}

A clear conclusion from the above tables is the fact that LKSVD is much better performing compared to the classic K-SVD, be it the universal dictionary approach or the image adaptive one. Indeed, the PSNR BSD68 results suggest that LKSVD is better than BM3D (by  $\sim0.5$dB) and WNNM (by $\sim0.1$dB) as well. Table \ref{table-Set12} zooms-in on this comparison, by comparing the various K-SVD versions (image-adaptive dictionary, universally trained dictionary, and the learned mode proposed in this work) on the images in set12. As can be seen, the learned scheme is consistently superior to its predecessors, apart from on the image \textsf{Barbara}, where the image adaptive dictionary seems to perform better. This is expected due to the unique textures and their quantities in this image, which our universal network is not accommodating properly.
As a final note we add that Table \ref{table-BSD-SSIM} shows that the ordering of the methods remains the same as we move from PSNR to the SSIM quality measure, which explains our choice to use PSNR for the rest of the experiments. 

We proceed by exploring the effect of $p$ (patch size), $m$ (dictionary size) and $K$ (number of denoising steps) on the LKSVD performance. We denote by $\text{LKSVD}_{K,p,m}$ the result for the proposed architecture with these specified parameters. Table \ref{table-compared-struct} presents the obtained results for  the two benchmarks (BSD68 and Set12) and a noise level of $\sigma=25$. As can be seen, even with $[p,m,K]=[64,256,1]$, LKSVD is markedly better than the classic K-SVD. As $K$ grows, the performance improves by $\sim0.1$dB per each additional denoising round. A boost in performance is also obtained when growing the patch-size to $16\times 16$ while preserving the redundancy factor of the dictionary. This also shows that the proposed scheme has the capacity to yield results that go beyond the ones reported in Tables \ref{table-BSD} and \ref{table-Set12}.

\begin{table*}[!t]
\centering
\scalebox{1}{
\begin{tabular}{ c | c | c c c c c c }
Dataset & Noise &  $\mbox{KSVD}_1$ & $\mbox{KSVD}_2$ & $\text{LKSVD}_{1,8,256}$   & $\text{LKSVD}_{3,8,256}$ & $\text{LKSVD}_{1,16,1024}$ &  $\text{LKSVD}_{2,16,1024}$  \\ 
\hline
BSD 68& \multirow{2}{*}{25} & 28.32 & 28.28 & 28.76 & 28.96 & 28.95 & 29.07  \\ 
Set 12 & & 29.41 & 29.17 & 29.76 & 30.05 & 30.09 & 30.22  \\  
\bottomrule
    \end{tabular}}
\space
\caption{LKSVD parameter effect: Denoising performance (PSNR [dB]) for $\sigma=25$ on BSD68 and Set12 while varying $p,m,K$.\label{table-compared-struct}}
\end{table*}

\begin{figure*}[!ht]
\centering
\begin{tabular}{ c  c  c  c  c  c  c}
True & Noisy & BM3D & WNNM & $\mbox{KSVD}_1$ & DnCNN & LKSVD \\ 
\hline
\vspace{-0.1in}\\
\includegraphics[width=0.11\textwidth, height=0.07\textwidth]{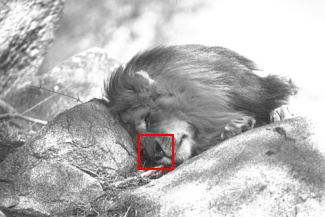} &
\includegraphics[width=0.11\textwidth, height=0.07\textwidth]{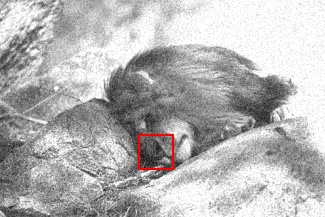}&
\includegraphics[width=0.11\textwidth, height=0.07\textwidth]{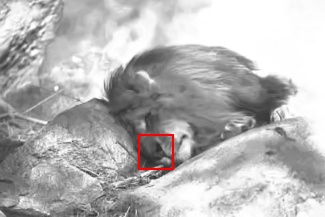} & \includegraphics[width=0.11\textwidth, height=0.07\textwidth]{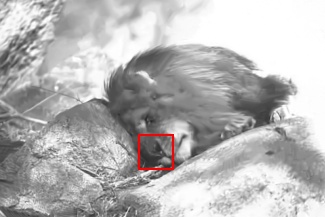} & \includegraphics[width=0.11\textwidth, height=0.07\textwidth]{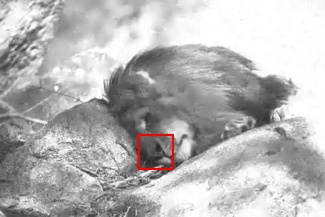} & \includegraphics[width=0.11\textwidth, height=0.07\textwidth]{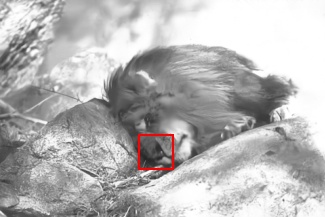} & \includegraphics[width=0.11\textwidth, height=0.07\textwidth]{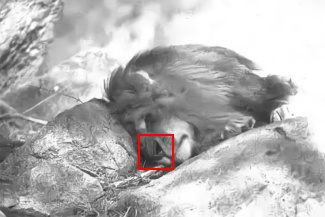} \\

\includegraphics[width=0.11\textwidth, height=0.11\textwidth]{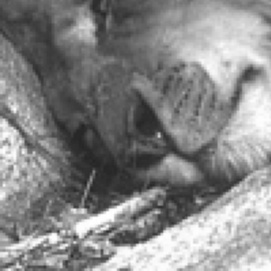} &
\includegraphics[width=0.11\textwidth, height=0.11\textwidth]{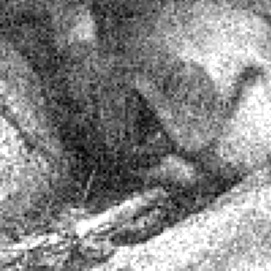}&
\includegraphics[width=0.11\textwidth, height=0.11\textwidth]{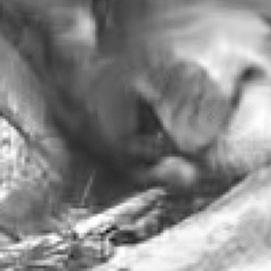} & \includegraphics[width=0.11\textwidth, height=0.11\textwidth]{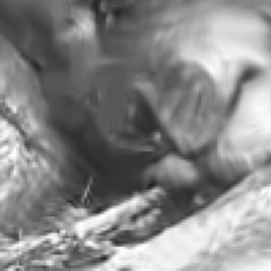} & \includegraphics[width=0.11\textwidth, height=0.11\textwidth]{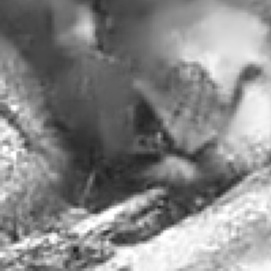} & \includegraphics[width=0.11\textwidth, height=0.11\textwidth]{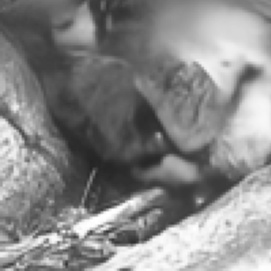} & \includegraphics[width=0.11\textwidth, height=0.11\textwidth]{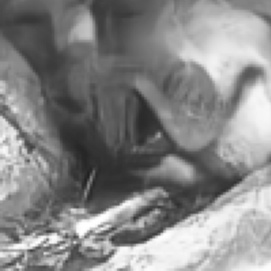} \\

 &
\includegraphics[width=0.11\textwidth, height=0.07\textwidth]{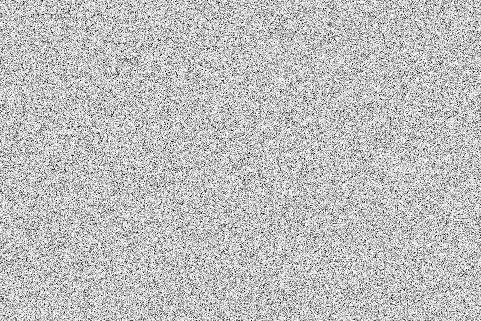}&
\includegraphics[width=0.11\textwidth, height=0.07\textwidth]{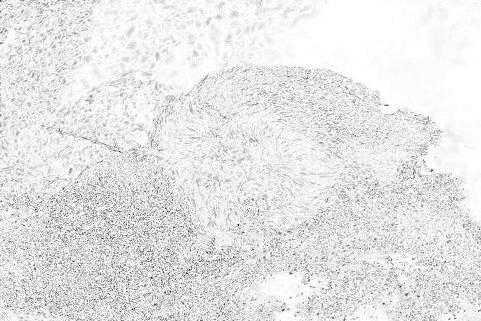} & \includegraphics[width=0.11\textwidth, height=0.07\textwidth]{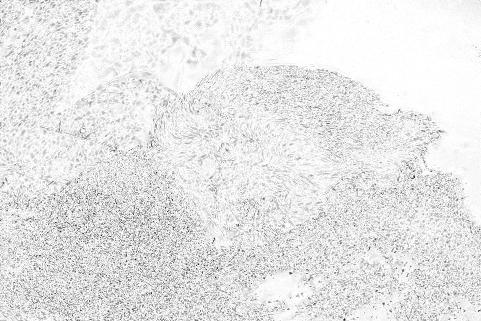} & \includegraphics[width=0.11\textwidth, height=0.07\textwidth]{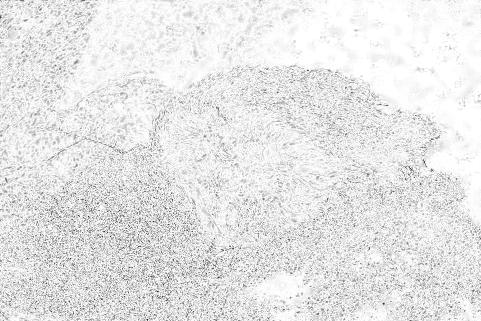} & \includegraphics[width=0.11\textwidth, height=0.07\textwidth]{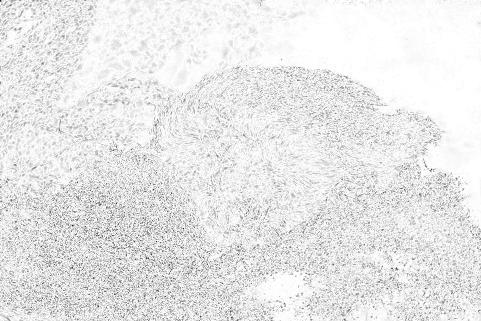} & \includegraphics[width=0.11\textwidth, height=0.07\textwidth]{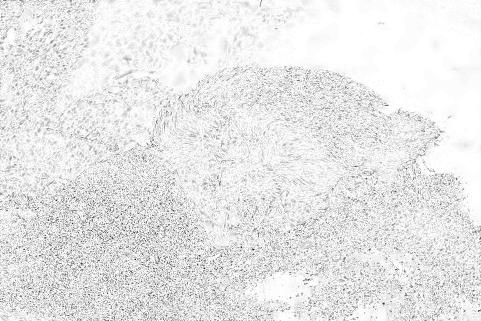} \\

& PSNR=20.86 & PSNR=27.07 & PSNR=27.26 & PSNR=27.20 & PSNR=27.67 & PSNR=27.47 \\

\end{tabular}
\caption{Denoising results for noise level $\sigma=25$.\label{fig-images25_0}}
\end{figure*}

\begin{figure*}[!ht]
\centering
\begin{tabular}{ c  c  c  c  c  c  c}
True & Noisy & BM3D & WNNM & $\mbox{KSVD}_1$ & DnCNN & LKSVD \\ 
\hline
\vspace{-0.1in}\\
\includegraphics[width=0.11\textwidth, height=0.07\textwidth]{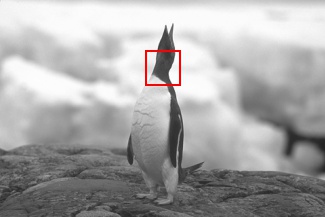} &
\includegraphics[width=0.11\textwidth, height=0.07\textwidth]{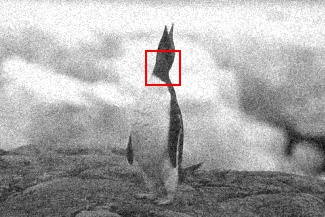}&
\includegraphics[width=0.11\textwidth, height=0.07\textwidth]{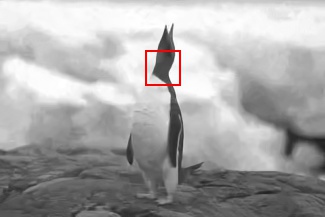} & \includegraphics[width=0.11\textwidth, height=0.07\textwidth]{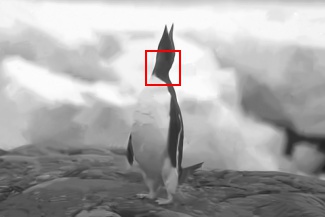} & \includegraphics[width=0.11\textwidth, height=0.07\textwidth]{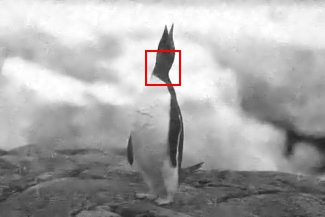} & \includegraphics[width=0.11\textwidth, height=0.07\textwidth]{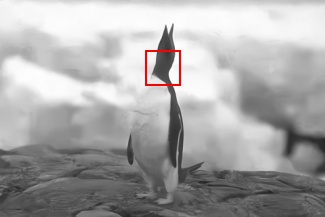} & \includegraphics[width=0.11\textwidth, height=0.07\textwidth]{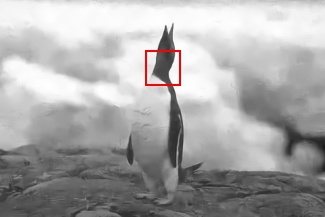} \\

\includegraphics[width=0.11\textwidth, height=0.11\textwidth]{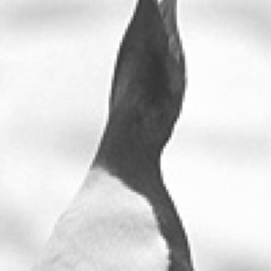} &
\includegraphics[width=0.11\textwidth, height=0.11\textwidth]{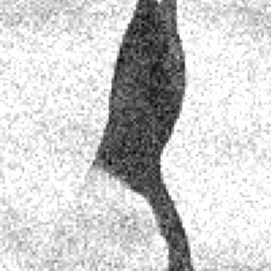}&
\includegraphics[width=0.11\textwidth, height=0.11\textwidth]{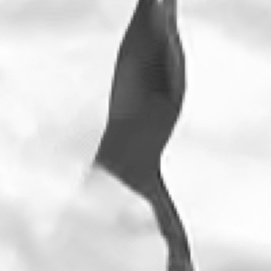} & \includegraphics[width=0.11\textwidth, height=0.11\textwidth]{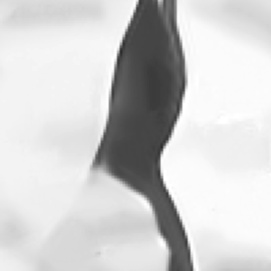} & \includegraphics[width=0.11\textwidth, height=0.11\textwidth]{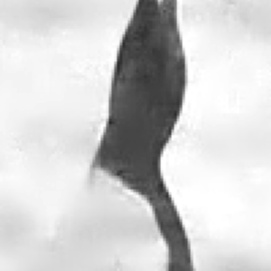} & \includegraphics[width=0.11\textwidth, height=0.11\textwidth]{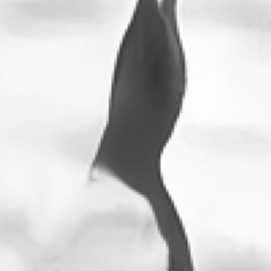} & \includegraphics[width=0.11\textwidth, height=0.11\textwidth]{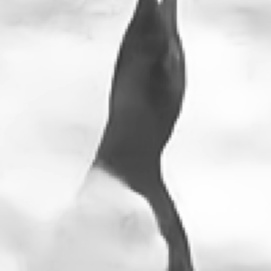} \\

 &
\includegraphics[width=0.11\textwidth, height=0.07\textwidth]{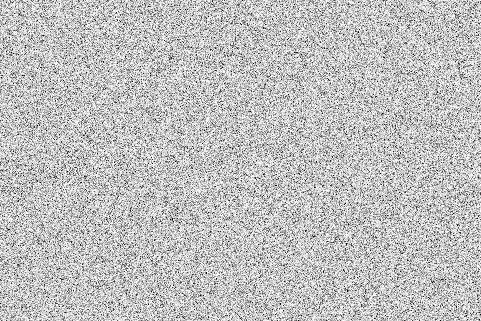}&
\includegraphics[width=0.11\textwidth, height=0.07\textwidth]{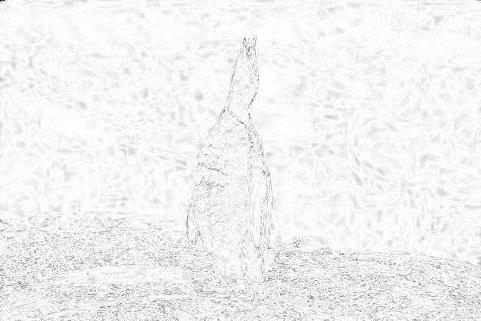} & \includegraphics[width=0.11\textwidth, height=0.07\textwidth]{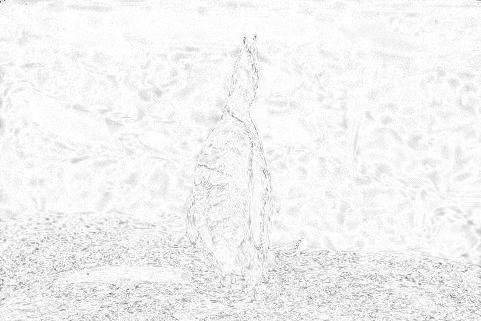} & \includegraphics[width=0.11\textwidth, height=0.07\textwidth]{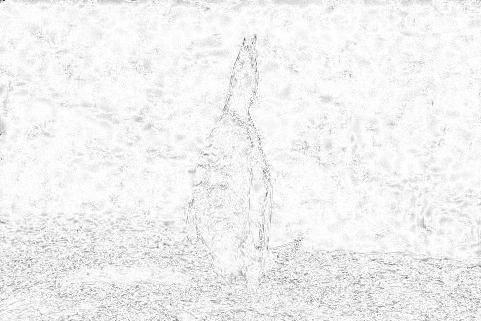} & \includegraphics[width=0.11\textwidth, height=0.07\textwidth]{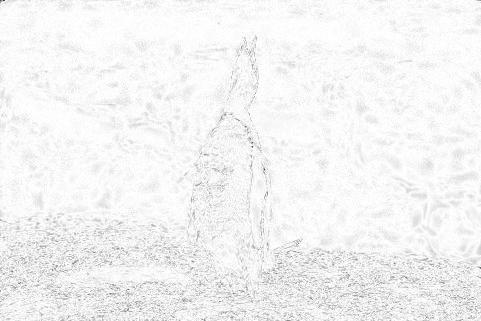} & \includegraphics[width=0.11\textwidth, height=0.07\textwidth]{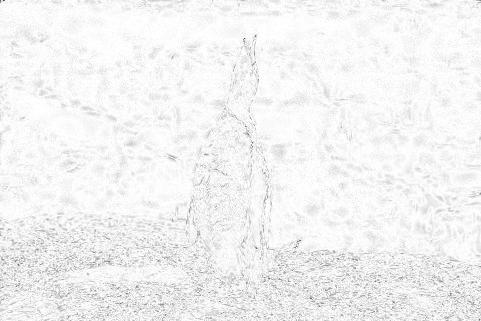} \\

& PSNR=20.67 & PSNR=32.42 & PSNR=32.59 & PSNR=31.79 & PSNR=33.02 & PSNR=32.61\\

\end{tabular}
\caption{Denoising results for noise level $\sigma=25$.\label{fig-images25_1}}
\end{figure*}

\begin{figure*}[!ht]
\centering
\begin{tabular}{ c  c  c  c  c  c  c}
True & Noisy & BM3D & WNNM & $\mbox{KSVD}_1$ & DnCNN & LKSVD \\ 
\hline
\vspace{-0.1in}\\
\includegraphics[width=0.11\textwidth, height=0.07\textwidth]{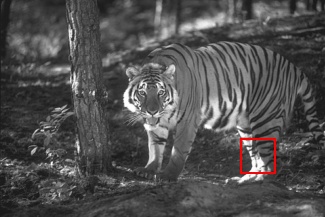} &
\includegraphics[width=0.11\textwidth, height=0.07\textwidth]{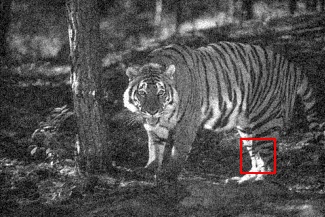}&
\includegraphics[width=0.11\textwidth, height=0.07\textwidth]{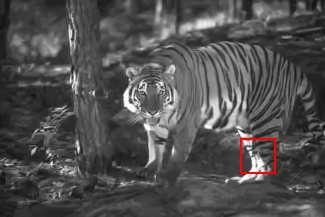} & \includegraphics[width=0.11\textwidth, height=0.07\textwidth]{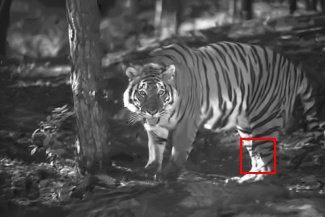} & \includegraphics[width=0.11\textwidth, height=0.07\textwidth]{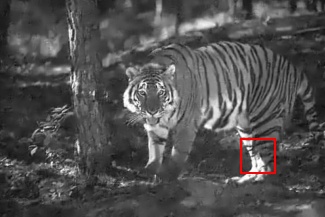} & \includegraphics[width=0.11\textwidth, height=0.07\textwidth]{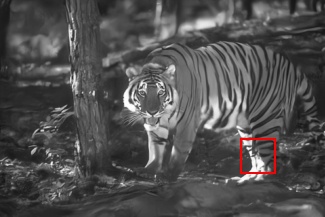} & \includegraphics[width=0.11\textwidth, height=0.07\textwidth]{KSVD_rec_2.jpg} \\

\includegraphics[width=0.11\textwidth, height=0.11\textwidth]{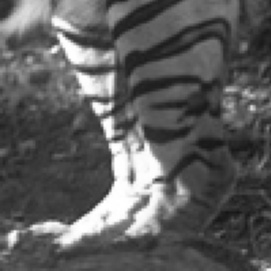} &
\includegraphics[width=0.11\textwidth, height=0.11\textwidth]{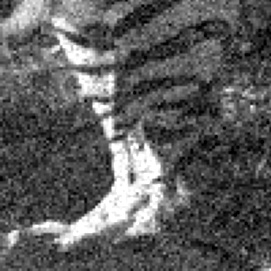}&
\includegraphics[width=0.11\textwidth, height=0.11\textwidth]{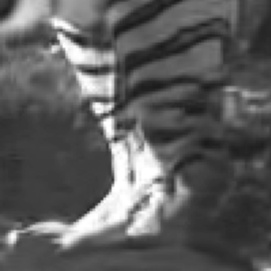} & \includegraphics[width=0.11\textwidth, height=0.11\textwidth]{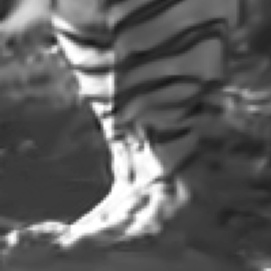} & \includegraphics[width=0.11\textwidth, height=0.11\textwidth]{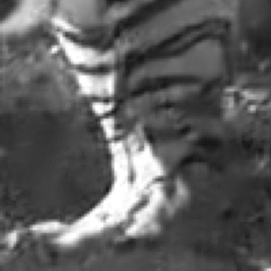} & \includegraphics[width=0.11\textwidth, height=0.11\textwidth]{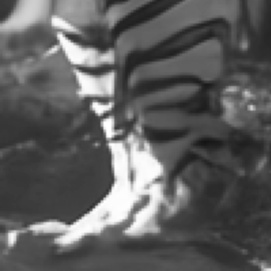} & \includegraphics[width=0.11\textwidth, height=0.11\textwidth]{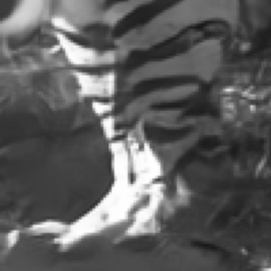} \\

 &
\includegraphics[width=0.11\textwidth, height=0.07\textwidth]{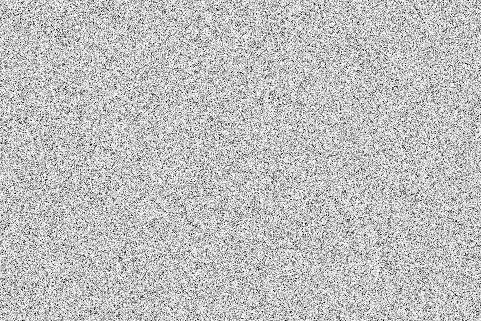}&
\includegraphics[width=0.11\textwidth, height=0.07\textwidth]{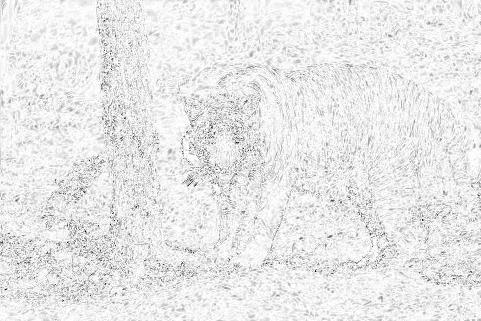} & \includegraphics[width=0.11\textwidth, height=0.07\textwidth]{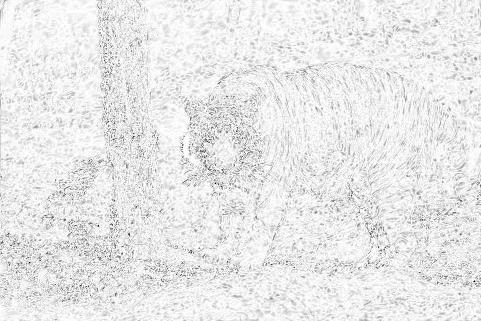} & \includegraphics[width=0.11\textwidth, height=0.07\textwidth]{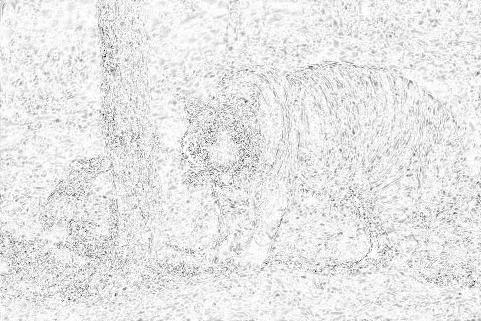} & \includegraphics[width=0.11\textwidth, height=0.07\textwidth]{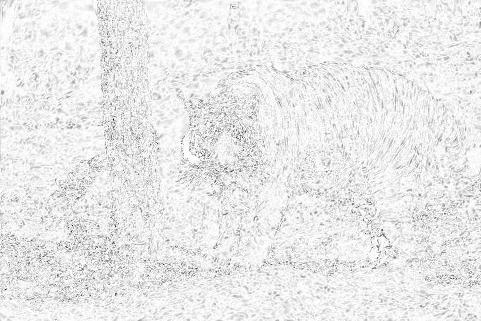} & \includegraphics[width=0.11\textwidth, height=0.07\textwidth]{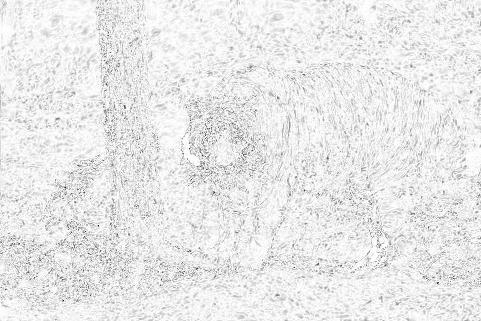} \\

 & PSNR=20.33 & PSNR=28.21 & PSNR=28.34 & PSNR=27.92 & PSNR=28.76 & PSNR = 28.54 \\

\end{tabular}
\caption{Denoising results for noise level $\sigma=25$.\label{fig-images25_2}}
\end{figure*}

We conclude by comparing the $\text{LKSVD}_{2,16,1024}$ with recent learning-based denoising competitors: TNRD \cite{chen2016trainable}, NLNet \cite{lefkimmiatis2017non}, DnCNN \cite{zhang2017beyond} and NLRNet \cite{liu2018non}. The results are shown in Table \ref{table-big-archi}, referring to the two benchmarks. As can be seen, our scheme surpasses TNRD \cite{chen2016trainable} and even the non-local deeply-learned denoiser by Lefkimmiatis \cite{lefkimmiatis2018universal, lefkimmiatis2017non}. Still, there is a gap between LKSVD and the best performing denoisers DnCNN \cite{zhang2017beyond} and NLRNet \cite{liu2018non}. 
We conclude this part by presenting visual results of the various methods compared. Figures \ref{fig-images25_0}, \ref{fig-images25_1}, and \ref{fig-images25_2} show the denoising results of BM3D, WNNM, $\mbox{KSVD}_1$, DnCNN, and LKSVD. 
All these figures refer to a noise level of $\sigma=25$ and the images used are taken from the BSD68 test set. For each figure, the top row presents the original image, the noisy one, and several cleaned images obtained by different methods. The middle row shows a zoomed in portion of the denoised images. The bottom row displays the absolute difference between the original and the cleaned images, where darker regions corresponds to locations where the images differ the most. As can be seen (both from the zoomed-in portions and the difference images), the visual quality of the LKSVD is somewhere in between WNNM and DnCNN,  exceeding BM3D and $\text{KSVD}_1$.


\subsection{The Network Inner Works}

\begin{figure}[!ht]
\centering
\begin{tabular}{ l} \vspace{-0.3in} \hspace{-0.8in}
\includegraphics[width=0.6\textwidth, height=0.45\textwidth]{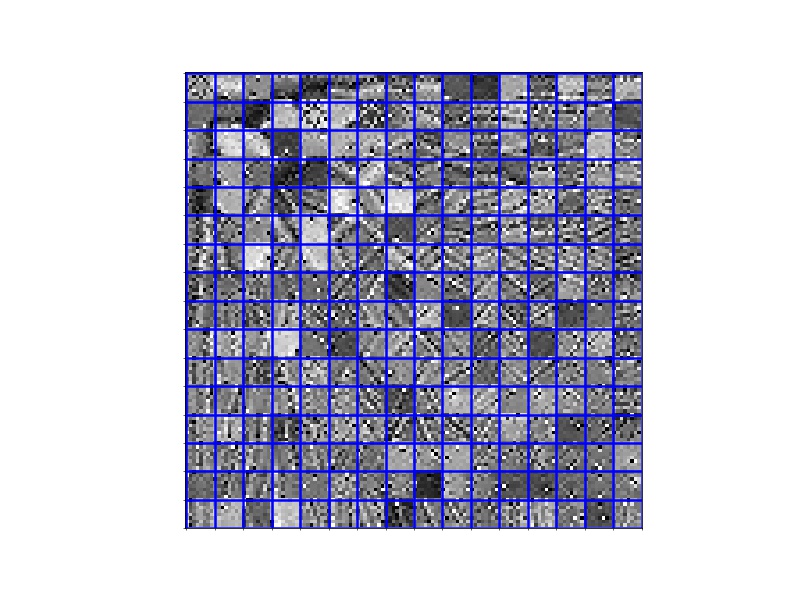} \\ \hspace{-0.8in}
\includegraphics[width=0.6\textwidth, height=0.45\textwidth]{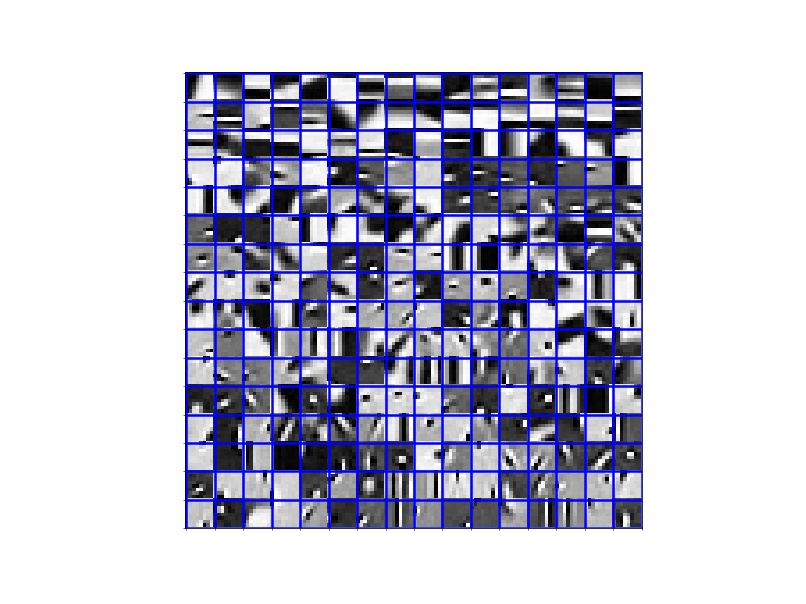}\\ 
\end{tabular}
\caption{Comparison of the dictionaries (top: LKSVD, bottom:  $\mbox{KSVD}_2$) for noise level $\sigma=25$.\label{fig-dic25}}
\end{figure}

\begin{figure}[!ht]
\centering
\begin{tabular}{ c}
\includegraphics[width=0.5\textwidth]{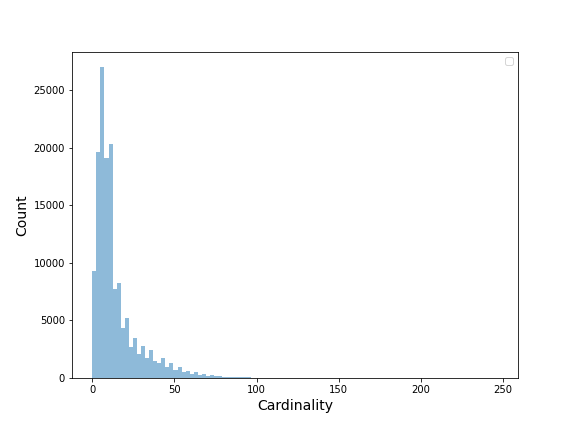} \\ \includegraphics[width=0.5\textwidth]{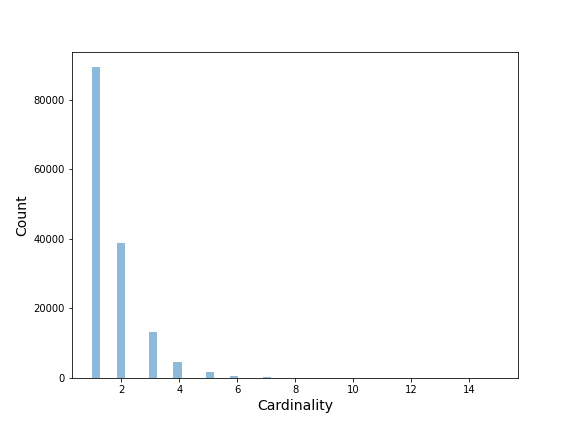}\\ 
\end{tabular}
\caption{Comparison of the cardinality histograms of the sparse code for noise level  $\sigma=25$ (top: LKSVD, bottom:  $\mbox{KSVD}_2$).\label{fig-cardinality25}}
\end{figure}

\begin{figure}[!ht]
\centering
\includegraphics[width=0.5\textwidth]{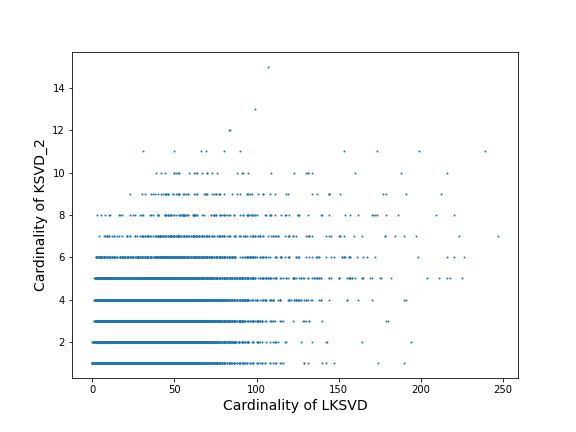}
\caption{The patch cardinality of the $\text{KSVD}_2$ as a function of the cardinality obtained by the LKSVD for randomly chosen patches.\label{fig-coupling-sparse}}
\end{figure}

We turn to have a closer look at the parameters learned by our proposed network and the data flowing in it. We do so by comparing the dictionary and the sparse codes obtained from LKSVD (referring to $\text{LKSVD}_{1,8,256}$) with its universal model-based method $\text{KSVD}_2$. Figure~\ref{fig-dic25} shows the learned dictionaries obtained from the $\text{KSVD}_2$~\cite{elad2006image}, and LKSVD trained on noisy images with $\sigma=25$. Figure~\ref{fig-cardinality25}
presents the distribution of the cardinalities of the sparse codes obtained in both $\text{KSVD}_2$ and LKSVD. Figure~\ref{fig-coupling-sparse} completes this comparison by showing the cardinality of the sparse codes obtained from the two methods one versus the other for each patch. 

As can be clearly seen from Figure~\ref{fig-dic25}, the LKSVD dictionary is markedly different from the universal one derived in \cite{elad2006image}. Indeed, the atoms learned by our scheme seem to be more local or edge-like, capturing very specific and small-scale patterns of the patches, while the atoms of the $\text{KSVD}_2$ method tend to be more global. Why would local atoms behave better than global ones? The answer is given in Figures~\ref{fig-cardinality25} and \ref{fig-coupling-sparse}. We see that the obtained sparse representations are in fact not sparse at all, and may contain number of non-zeros that could even exceed the dimension of the patches being treated. This implies that the LKSVD gathers many more atoms for representing each patch, thus being able to operate with more local atoms.

A natural question to pose is whether our new scheme contradicts sparse modeling altogether? Surprisingly, the answer is negative -- dense representations do make sense even if the original signals emerge as sparse compositions of atoms from a given dictionary. This is in-fact the outcome of the Minimum-Mean-Squared-Error (MMSE) estimation, as described in \cite{mmse2009}. Thus, as our end-to-end training loss is the MSE, it should not come as a surprise that the pursuit obtained by our unfolded LISTA tunes itself to this behavior. Indeed, the same phenomenon can be expected with LISTA in general~\cite{gregor2010learning}.

\subsection{Complexity}

\begin{figure}[!ht]
\centering
\includegraphics[width=0.5\textwidth]{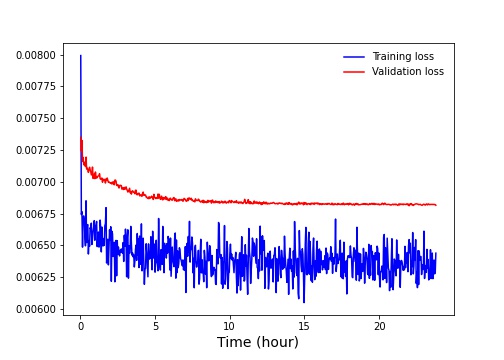}
\caption{Training and Validation losses of our proposed method for $\sigma=25$.\label{training-time}}
\end{figure}

Tables \ref{table-complex} and \ref{table-run-time} shed more light on the above results by presenting respectively the model complexities and the run times involved in this experiment. As can be seen, our network LKSVD uses about 10\% of the overall number of parameters compared to the better performing methods. More precisely, the total number of algebraic operations for inference is $\mathcal{O}(pmTKN)$ where $N$ is the image size (e.g. $256^2$), $p$ is the patch-size (e.g. $64$), $m$ is the number of atoms (e.g. $256$), $T$ is the number of unfoldings (e.g. $10$), and $K$ is the number of EPLL rounds (e.g. $3$. 

In terms of run-time, the proposed method is very competitive with respect to the model based methods, while being on par when compared to DnCNN. We should note that our GPU implementation could have have been much faster if all the pursuit would be merged to a common process, something that implies that the multiplication by each atom is done on all patches together, forming a convolution. Put in other words, each iteration in LISTA for all the patches can be represented as $m$ convolutions. Figure~\ref{training-time} shows the training and validation losses, demonstrating that our proposed method could be fully trained in less than $10$ hours.

\begin{table}[!h]
\centering
\scalebox{0.8}{
\begin{tabular}{ c | c | c c c c  c }
Dataset & Noise & TNRD &  NLNet & DnCNN & NLRNet & $\text{LKSVD}_{2,16,1024}$ \\ 
\hline
\multirow{3}{*}{BSD 68}& 15  & 31.42 & 31.52 & \textbf{31.73} & \textbf{31.88} & 31.54 \\  
& 25 & 28.92 & 29.03 & \textbf{29.23} & \textbf{29.41} & 29.07 \\  
& 50 & 25.97 & 26.07 & \textbf{26.23} & \textbf{26.47} &  26.13 \\
\hline
\multirow{3}{*}{Set 12}& 15 & 32.50 & - & \textbf{32.86} & \textbf{33.16} & 32.61 \\  
& 25 & 30.06 & - & \textbf{30.44} & \textbf{30.80} & 30.22  \\  
& 50 & 26.81  & - &  \textbf{27.18}  & \textbf{27.64} &   27.04  \\
\bottomrule
\end{tabular}}
\space
\caption{LKSVD versus learned methods: Denoising performance (PSNR [dB]) for various noise levels on BSD68 and Set12. Results exceeding LKSVD are marked in bold. \label{table-big-archi}}
\end{table}

\begin{table}[!h]
\centering
\scalebox{1}{
\begin{tabular}{ c | c | c | c }
 & DnCNN & NLRNet & LKSVD \\ 
\hline
Max effective depth & 17  & 38 & 21 \\  
\hline
Parameter sharing & No &  Yes & Yes  \\  
\hline
Parameter no. & 554k &  330k & 45k \\
\bottomrule
\end{tabular}}
\space
\caption{Model complexities comparison of our proposed scheme and two  state-of-the-art networks. \label{table-complex}}
\end{table}

\begin{table*}[!t]
\centering
\scalebox{1}{
\begin{tabular}{ c | c c c c c c}
Image size & BM3D & WNNM & $\mbox{KSVD}_2$ & DNCNN &$\text{LKSVD}_{1,8,256}$  & $\text{LKSVD}_{3,8,256}$  \\ 
\hline
$256\times 256$ &  0.65 & 203.1 & 65.25 & 0.86 / 0.0045 & 0.91 / 0.033  & 2.73 / 0.10 \\ 
$512\times 512$ & 2.85 & 773.2 & 261.4 & 3.86 / 0.0059 & 4.06 / 0.13 & 13.7 / 0.41  \\  
\bottomrule
    \end{tabular}}
\space
\caption{Run time (in seconds) of different methods on images of size $256\times 256$ and $512 \times 512$. We give the run times on CPU (left) and GPU (right) for DnCNN and the LKSVD.\label{table-run-time}}
\end{table*}

\begin{figure*}[!ht]
\centering
\begin{tabular}{ c  c  c  c  c  c  c}
True & Noisy & BM3D & WNNM & $\mbox{KSVD}_1$ & DnCNN & LKSVD \\ 
\hline
\vspace{-0.1in}\\
\includegraphics[width=0.11\textwidth, height=0.07\textwidth]{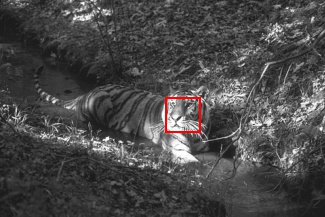} &
\includegraphics[width=0.11\textwidth, height=0.07\textwidth]{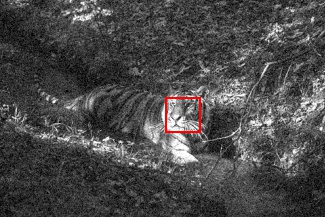}&
\includegraphics[width=0.11\textwidth, height=0.07\textwidth]{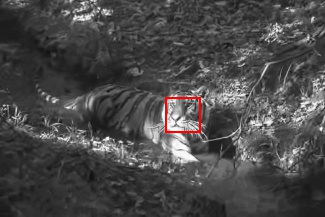} & \includegraphics[width=0.11\textwidth, height=0.07\textwidth]{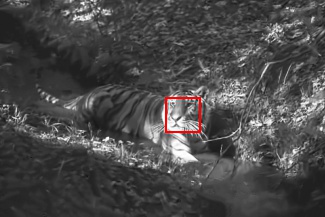} & \includegraphics[width=0.11\textwidth, height=0.07\textwidth]{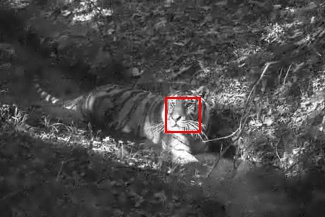} & \includegraphics[width=0.11\textwidth, height=0.07\textwidth]{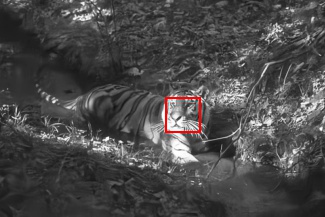}
& \includegraphics[width=0.11\textwidth, height=0.07\textwidth]{KSVD_rec_3.jpg} \\

\includegraphics[width=0.11\textwidth, height=0.11\textwidth]{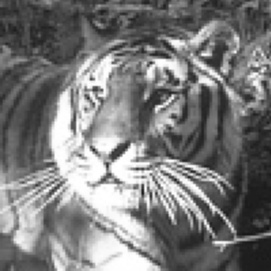} &
\includegraphics[width=0.11\textwidth, height=0.11\textwidth]{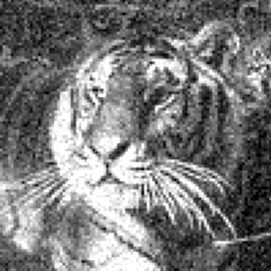}&
\includegraphics[width=0.11\textwidth, height=0.11\textwidth]{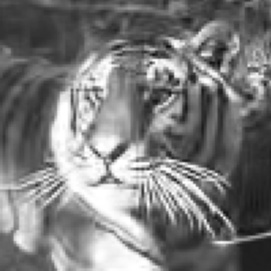} & \includegraphics[width=0.11\textwidth, height=0.11\textwidth]{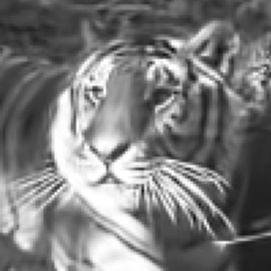} & \includegraphics[width=0.11\textwidth, height=0.11\textwidth]{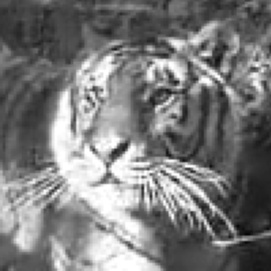} & \includegraphics[width=0.11\textwidth, height=0.11\textwidth]{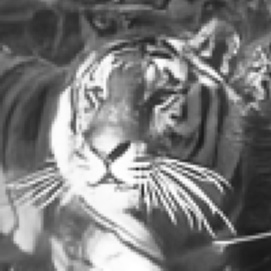} & \includegraphics[width=0.11\textwidth, height=0.11\textwidth]{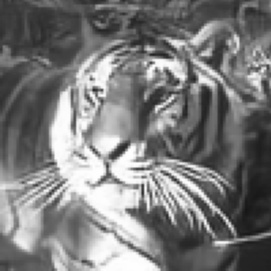} \\

 &
\includegraphics[width=0.11\textwidth, height=0.07\textwidth]{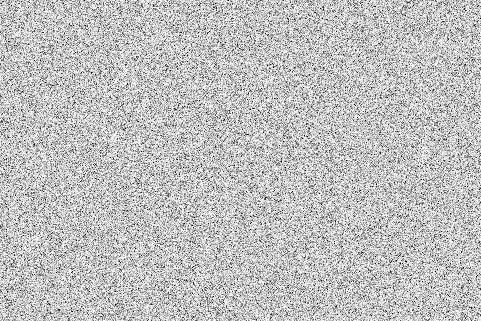}&
\includegraphics[width=0.11\textwidth, height=0.07\textwidth]{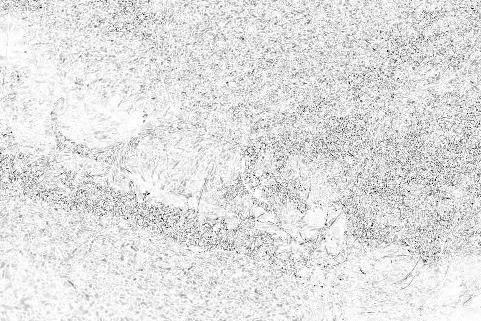} & \includegraphics[width=0.11\textwidth, height=0.07\textwidth]{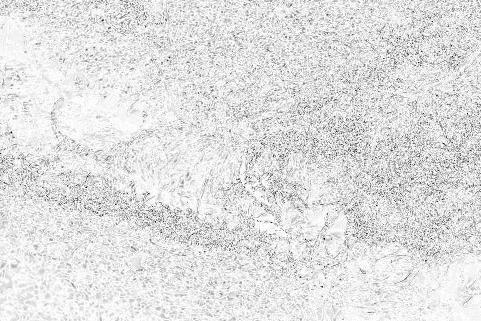} & \includegraphics[width=0.11\textwidth, height=0.07\textwidth]{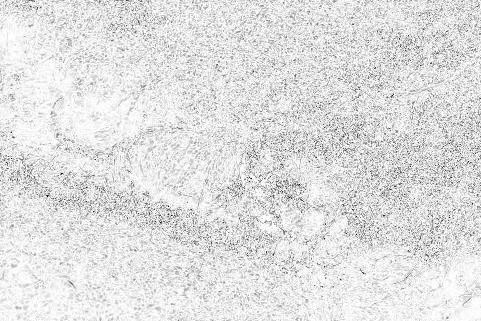} & \includegraphics[width=0.11\textwidth, height=0.07\textwidth]{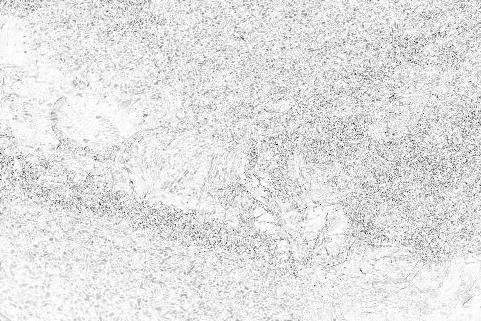} & \includegraphics[width=0.11\textwidth, height=0.07\textwidth]{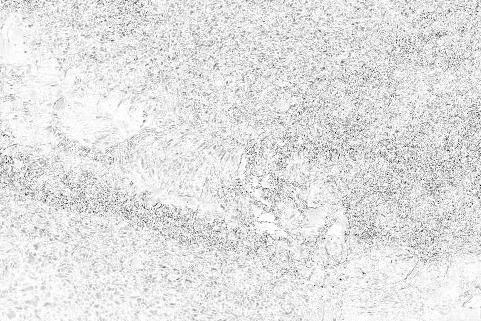} \\
& PSNR=20.39 & PSNR=25.60 & PSNR=25.86 & PSNR=25.76 & PSNR=26.27 & PSNR=26.16\\ 

\end{tabular}
\caption{Denoising results for noise level $\sigma=25$.\label{fig-images25_3}}
\end{figure*}

\begin{figure*}[!ht]
\centering
\begin{tabular}{ c  c  c  c  c}
True & Noisy & WNMM &  LKSVD & LSKVD-U \\ 
\hline
\vspace{-0.1in}\\
\includegraphics[width=0.15\textwidth, height=0.15\textwidth]{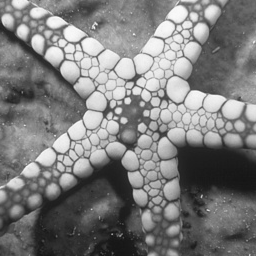} &
\includegraphics[width=0.15\textwidth, height=0.15\textwidth]{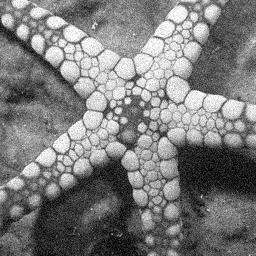} &
\includegraphics[width=0.15\textwidth, height=0.15\textwidth]{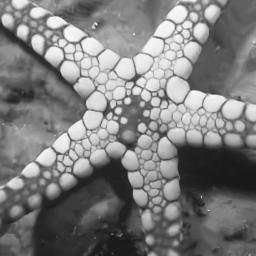}&
\includegraphics[width=0.15\textwidth, height=0.15\textwidth]{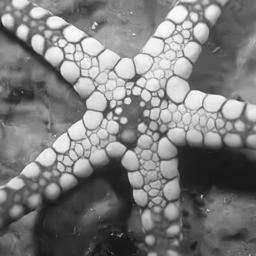} & \includegraphics[width=0.15\textwidth, height=0.15\textwidth]{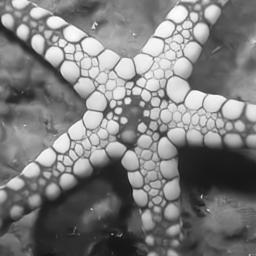} \\
 & PSNR=24.61 & PSNR=31.82 & PSNR=31.78 & PSNR=32.04\\
\end{tabular}
\caption{Self-adaptation: Denoising results for the image \textsf{Starfish} with noise level $\sigma=15$. As can be seen, the additional adaptation leads to $0.26$dB boost in performance, surpassing the WNNM method. \label{Starfish_Unsupervised}}
\end{figure*}

%% file: 5-Conclu.tex
\section{Discussion and Conclusions}
\label{section-Discussion}

Why should we bring classical methods and deep learning together? Clearly, the performance of the proposed method still falls short when compared to leading state-of-the-art deep networks. Below we dive deeper into this question, discussing the motivation behind this work, and explaining the merits and prospects of the specific network proposed. 

\subsection{Why bother improving K-SVD denoising?}

The rationale behind this work goes beyond a simple improvement of the K-SVD denoising algorithm. Indeed, our motivation is drawn from the hope to propose systematic ways of designing deep-learning architectures and connecting novel solutions to classical algorithms. 

A fundamental question nowadays in computational imaging is whether old/classic methods should be discarded and replaced by their deep-learning alternatives. In the context of image denoising, classical methods focused on data modeling and optimization, and searched for ways to identify and exploit the redundancies existing in the visual data. The recent deep networks, which lead the denoising performance charts today, take an entirely different route, targeting the inference stage directly, and learning their parameters for optimized end-to-end performance. Now that these methods are getting close to touch their ceiling, our work comes to argue that the classic methods are still very much relevant, and could become key in breaking such barriers. We believe that classic image processing algorithms will have a comeback for this exact reason. 

Adopting a different point of view, this work offers a migration from intuitively chosen architectures, as many recent papers have offered, towards well-justified ones based on domain knowledge of the problem  we are trying to solve. That is, the structure of the denoising problem is embedded into the deep learning architecture, making the overall algorithm enjoy both the flexibility of the deep methods, and the structure brought by the more classical approaches. The option of piling convolutions, ReLU's, batch-normalization steps, skip connections, strides and pooling operations, dilated filtering, and many other tricks, and seeking for best performing architectures by trial an error, has been the dominating approach so far. 

It is time to return to the theoretical foundations of signal and image processing in order to go beyond this point.
Relying on sparse representation modeling, the K-SVD network we introduce in this work has a clear objective, a concise structure, and yet it works quite well. We believe that the results shown here stand as yet another testimony for the central role that sparse modeling plays in broad data processing. We hope that a fusion of the past knowledge with the new deep-learning view could bring us to the next levels in a long list of applications in image processing. This paper offers a valuable step in this direction, as the architecture we propose offers a clear interpretability of its features (being the sparse representations of overlapping image patches) and parameters (being the learned dictionary and the threshold for the pursuit).

And related to the above, here is an interesting question: What is the simplest possible network, in terms of the number of free parameters to learn and the number of computations to apply, for getting state-of-the-art image denoising? In single-image super-resolution it has become common practice in the literature to compare different solutions by considering their complexity as well (e.g., \cite{Timofte_2016_CVPR}). This is done by showing points in a 2D graph of PSNR versus computational cost. Doing the same in image denoising may reveal interesting patterns. The general deep-learning based methods, while showing the best PSNR, tend to be quite heavy and cumbersome. Could much lighter networks perform nearly as well (and perhaps even better)? In this work we offer one such avenue to explore, and we are certain that many others will follow. 

\subsection{Going Beyond Sparsity?}

Why has it been so easy to outperform the original K-SVD denoising algorithm in the first place? A possible answer could be that this algorithm builds its cleaning abilities on two prime forces: (i) the spatial redundancy that exists in image patches, exposed by the sparse modeling; and (ii) the patch-averaging effect, which has an MMSE flavor to it \cite{papyan2015multi}. Many of the better performing competitors strengthen their performance by considering several additional ideas:
\begin{itemize}
    
    \item {\bf Non-Locality:} Non-local self-similarity can be practiced as an additional prior, as done by BM3D \cite{dabov2007video} and low-rank modeling \cite{gu2014weighted, Yair_2018_CVPR}. Indeed, the paper by Mairal et. al \cite{mairal2009non} extended the K-SVD denoising by incorporating joint sparsity on groups of patches, this way introducing non-locality. Broadly speaking, non-local methods are known to be effective in capturing the correlation between far-apart patches, leading to improved restoration. 
    
    \item{\bf Patch Consensus:} Patch based methods must address the disagreement found between overlapping patches. The original K-SVD    scheme we embark from in this paper proposed an averaging\footnote{While the original K-SVD denoising algorithm has used a plain averaging, we deploy a slightly improved weighted option, due to its simplicity in the context of a learned machine.} of these patches. However, the EPLL approach \cite{zoran2011learning} suggests a far better strategy, by imposing the prior on patches taken from the resulting image, rather than ones extracted from the measured one. In the context of sparse modeling, this idea boils down to an iterated K-SVD algorithm, as was shown in \cite{sulam2015expected}. In such a scheme the cleaned image is aggregated and broken to patches again for subsequent pursuit. We have deployed this very idea in an elementary way by replicating the filtering process. Closely related alternatives to this strategy are the SOS boosting method \cite{romano2015boosting} and the deployment of the CSC model \cite{papyan2017convolutional}. 
    
    \item {\bf Multi-Scale:} Multi-scale analysis of visual data seems to be a natural strategy to follow, and various papers have shown the benefit of this for image denoising \cite{papyan2015multi}. More specifically, a multi-scale extension of the K-SVD denoising algorithm has been considered in various practical ways \cite{sulam2014image,ophir2011multi,mairal2007multiscale,mairal2008learning}. 
\end{itemize}

\noindent The above suggests that K-SVD denoising in its original form carries a built-in weakness in it. Yet, the results in this paper suggest otherwise. Consider the more recent and better performing deep-learning based solutions. These alternatives seem to disregard these extra forces (at least explicitly), concentrating instead on capturing image intrinsic properties by a direct supervised learning of the inference process. Recent such convolutional neural networks (CNNs) for image restoration \cite{zhang2017beyond, mao2016image} achieve impressive performance over classical approaches. Do these methods exploit self-similarity? anything reminiscent of patch-consensus? a multi-scale architecture? One may argue that the answer is, at most, only partially positive, hidden by the wide receptive field and the global treatment that these networks entertain. Note that there are deep learning methods that explicitly use self-similarity in their processing  \cite{lefkimmiatis2017non,liu2018non}, however those do not necessarily improve over the simpler alternatives.

The conclusion we draw from the above is that there is room for introducing non-locality, patch-consensus and a multi-scale structure into the proposed K-SVD scheme, thereby driving the revised architecture towards even better results. Indeed, nothing is sacred in the K-SVD computational path, and the same treatment as done in this work could be given to well-performing classical denoising algorithms, such as BM3D \cite{dabov2007video}, kernel-based methods \cite{takeda2006kernel} and WNNM \cite{gu2014weighted}. We leave these ideas for future work.


\subsection{Unsupervised version of this architecture? }

This is perhaps a good time to recall that the denoising work in \cite{elad2006image} offered two strategies for getting the dictionary -- a globally universal approach that trains the dictionary off-line, and an image-adaptive alternative that trains on the noisy image patches themselves. Interestingly, despite the fact that the later (image-adaptive) approach was found to be better performing, the solution we put forward in this paper aligns solely with the first approach. Why? because the supervised strategy we adopt naturally leads to a single architecture that serves all images via the same set of parameters. Could we offer an unsupervised alternative, more in line with the image adaptive path? The answer, while tricky, is certainly be positive. A related approach of great relevance is \cite{ulyanov2018deep}, in which a chosen network architecture is trained on each image all over again. A similar concept could be envisioned, where our own K-SVD architecture is used for synthesizing the clean image. However, this raises some difficulties and challenges, which is why we leave this activity for future work.

Another, more immediate direction to answer the above challenge, is self-adaptation along the lines described in \cite{LIDIA}. The core idea is to run the trained universal network to create an initial denoised result, followed by an adaptation round to the incoming image by few epochs on the image itself and its initially cleaned version. Figure \ref{Starfish_Unsupervised} presents an example result of this idea on the image \textsf{Starfish}, showing a boost of $0.26$dB in its denoising. This adaptation is obtained via the exact same training procedure as the one used to train the LKSVD network, where the target image is replaced by the restored image obtained from the universally trained LKSVD network, and using only few minutes of training. 


\section{Conclusion}

This work shows that the good old K-SVD denoising algorithm \cite{elad2006image} can have a comeback and become much better performing, getting closer to leading deep-learning based denoisers. This is achieved very simply by setting its parameters in a supervised fashion, while preserving its exact original form. Our work have shown how to turn the K-SVD denoiser into a learnable architecture that enables back-propagation, and demonstrated the achieved boost in denoising performance. As the  discussion above reveals, our story goes beyond the K-SVD denoising and its improvement, towards more fundamental questions related to the role of deep-learning in contemporary image processing.